%% file: main.tex
\documentclass{article}
\usepackage{spconf,amsmath,graphicx,hyperref}
\usepackage{multicol,multirow}
\usepackage{booktabs} 
\usepackage{graphicx} 
\usepackage{xcolor}
\usepackage{booktabs}
\usepackage{caption} 

\title{StressTransfer: Stress-Aware Speech-to-Speech Translation \\with Emphasis Preservation}
\name{Xi Chen$^{1}$, Yuchen Song$^{1}$, Satoshi Nakamura$^{1, 2, \star}$
\thanks{$^{\star}$ indicates the corresponding author.}
}
\address{$^{1}$ The Chinese University of Hong Kong, China
$^{2}$ Nara Institute of Science and Technology, Japan}
\begin{document}
\ninept
\maketitle
\vspace*{-1cm}
\begin{abstract}
We propose a stress-aware speech-to-speech translation (S2ST) system that preserves word-level emphasis by leveraging LLMs for cross-lingual emphasis conversion. Our method translates source-language stress into target-language tags that guide a controllable TTS model. To overcome data scarcity, we developed a pipeline to automatically generate aligned training data and introduce the "LLM-as-Judge" 
for evaluation. Experiments show our approach substantially outperforms baselines in preserving emphasis while maintaining comparable translation quality, speaker intent, and naturalness. Our work highlights the importance of prosody in translation and provides an effective, data-efficient solution for preserving paralinguistic cues in S2ST.
\end{abstract}

\input{sec/sec1_introduction}
\input{sec/sec3_dataset}

\input{sec/sec4_stress_trans}
\input{sec/sec5_experienments}

\input{sec/sec6_evaluation}

\section{Conclusion}

In this work, we present StressTransfer, an emphasis-preserving speech-to-speech translation system that integrates LLM and prosody-controllable TTS backend. This addresses the gap in cross-lingual prosodic preservation for natural communication. We build an automatic Emphasis-Aligned speech-to-text translation dataset called EmphST-Instruct and conduct evaluations on EmphST-Bench, which show superior stress accuracy over baselines like GPT-4o-audio and Gemini-2.5-Pro, with competitive semantic quality on CoVoST-2. The subjective tests further demonstrate better emphasis transfer and fidelity. The ablation study confirm the value of EmphST-Instruct data and LLM-as-a-Judge reliability. 
As an initial effort to harness large models for stress-aware S2ST, StressTransfer serves as a simple yet effective baseline, paving the way for future exploration.


\vfill\pagebreak


\bibliographystyle{IEEEbib}
\bibliography{refs}

\end{document}

%% file: sec/sec1_introduction.tex
\section{Introduction}
Speech-to-speech Translation (S2ST) is a transformative technology for bridging global communication gaps. While early systems prioritized semantic accuracy, effective communication also relies on paralinguistic information like emphatic stress, which conveys focus, intent, and nuance~\cite{schuller2013computational}. Neglecting such cues can lead to misinterpretation and a loss of expressive impact.

The field has evolved significantly from traditional cascaded pipelines, comprising separate Automatic Speech Recognition (ASR) and Machine Translation (MT) modules~\cite{casacuberta2008recent, sperber2017neural}, to end-to-end (E2E) models that translate speech directly. Modern approaches leverage powerful pre-trained components, such as Whisper~\cite{radford2023robust} for speech encoding and Large Language Models (LLMs) for text decoding~\cite{huang2023speech,wu2023decoder}, to improve robustness. In parallel, S2ST has shifted from direct spectrogram prediction~\cite{jia2019direct,jia2022translatotron} to generating discrete speech units~\cite{lee2021direct, Inaguma2022UnitYTD}, a more stable approach exemplified by state-of-the-art multilingual systems like SeamlessM4T~\cite{barrault2023seamlessm4t,barrault2023seamless}, which excel at semantic translation.

However, a critical gap remains in preserving expressivity. While recent models like SeamlessExpressive~\cite{gong2024seamlessexpressivelm}, AudioPaLM~\cite{rubenstein2023audiopalm} and TransVIP~\cite{le2024transvip} can transfer general prosody and speaker timbre, the explicit preservation of lexical or sentential emphasis is largely unaddressed. Pioneering work by Do et al.~\cite{do2017preserving, do2018sequence} demonstrated the feasibility of jointly translating words and emphasis levels but also highlighted a major bottleneck: the dependence on small-scale, manually annotated datasets, which severely limits scalability and generalization.

To address this critical data bottleneck, we introduce EmphST-Instruct, a fully automated pipeline to synthesize large-scale, emphasis-aligned parallel corpora. Using this data, we develop an end-to-end Speech-to-text Translation (S2TT) model that ingests source audio and outputs target sentences interleaved with explicit emphasis markers. To foster rigorous evaluation, we also introduce EmphST-Bench, the first benchmark dedicated to assessing emphasis preservation in S2TT. Finally, we demonstrate how our emphasis-aware S2TT model can be incorporated into a cascaded S2ST system by pairing it with a controllable Text-to-speech synthesizer~\cite{du2024cosyvoice} that renders the predicted emphasis markers as prosodic prominence.

Through comprehensive experiments, we show that our system outperforms existing baselines~\cite{barrault2023seamlessm4t, barrault2023seamless, comanici2025gemini} in preserving speaker intent and expressive stress. In summary, our contributions are:

\noindent$\bullet$ We propose \textbf{EmphST-Instruct}, a fully automated pipeline for generating large-scale, emphasis-aligned parallel corpora, enabling scalable training of emphasis-aware translation models. In addition, we introduce \textbf{EmphST-Bench}, which is specifically designed to evaluate emphasis preservation in speech translation, featuring diverse stress patterns and automatic evaluation metrics.

\noindent$\bullet$ We develop an end-to-end Speech-to-text translation model that explicitly preserves lexical emphasis by predicting target-language text interleaved with stress markers directly from source speech.

\noindent$\bullet$ We construct a complete S2ST system by integrating our emphasis-preserving S2TT model with a controllable TTS synthesizer. Through comprehensive objective and subjective evaluations on both translated text and speech, we demonstrate that our system achieves state-of-the-art performance in preserving expressive stress across translation.

This work represents an initial attempt at stress-aware Speech-to-Speech Translation. We hope that the proposed data techniques, benchmark, and models will pave the way for future exploration in expressive speech translation.

%% file: sec/sec3_dataset.tex
\section{EmphST-Instruct}
\label{sec:dataset}

The scarcity of parallel speech-text data with cross-lingual emphasis alignment ($|\mathcal{D}_{\text{stress}}| \ll |\mathcal{D}_{\text{std}}|$) poses a significant bottleneck for training systems that preserve emphasis in speech translation. While techniques like controlled text-to-speech (TTS) can generate high-quality monolingual emphasis data~\cite{yosha2025whistress, yosha2025stresstest}, extending this to cross-lingual scenarios requires sophisticated semantic understanding to project emphasis markers, as stress realization lacks linguistic universals across languages.
\begin{figure}[t!]
    \centering
        \includegraphics[width=0.47\textwidth]{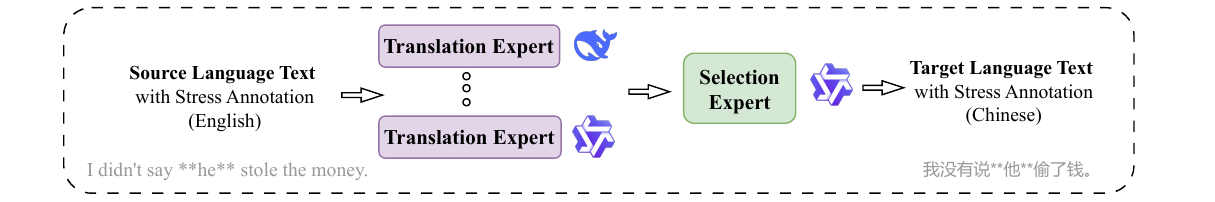} 
    \caption{Framework of the StressTransfer model.} 
    \label{fig:data_pipe} 
\end{figure}

To address this challenge, we introduce an innovative pipeline that leverages large language models (LLMs) for \textbf{LLM-assisted data construction}, generating a high-quality, emphasis-aligned S2TT dataset—\textbf{EmphST-Instruct}—from the Stress17k~\cite{yosha2025stresstest} and TinyStress~\cite{yosha2025whistress} corpus, an English resource containing speech-text pairs annotated with lexical emphasis. 
Our method translates the source English text to the target language (Chinese in this work) while preserving emphasis annotations, ensuring semantic fidelity and natural stress placement in the target language. This is achieved through a two-stage process for the training set: (1) multi-LLM translation for diverse candidates, and (2) quality assessment and selection using an additional LLM. By ensembling multiple LLMs, we mitigate individual model biases and enhance robustness, representing a novel approach to scalable, low-cost dataset construction for emphasis-preserving tasks. To ensure the quality, we involve human experts to check the sampled synthetic data and iteratively update the prompt for the selection expert model.

%

As shown in Figure.~\ref{fig:data_pipe}, in the first stage, we deploy LLMs as translation experts that generate target-language text with corresponding stress patterns given source text and stress annotations from Stress-17k~\cite{yosha2025stresstest} and TinyStress~\cite{yosha2025whistress}. To ensure translation diversity while maintaining quality and stress accuracy, we employ multiple LLM translation experts. In the second stage, an LLM serves as a selection expert to identify the optimal translation from the experts' outputs.


%% file: sec/sec4_stress_trans.tex
\section{Emphasis-preserving S2ST System}


\subsection{Overview}

Our emphasis-preserving speech-to-speech translation (S2ST) system operates as a cascaded pipeline, seamlessly integrating an end-to-end emphasis-preserving speech-to-text translation (S2TT) model with a controllable text-to-speech (TTS) synthesizer. The S2TT component processes input source speech to produce target-language text augmented with explicit emphasis markers, while the TTS module leverages these markers to synthesize speech with natural and intent-preserving prosody. Inspired by the LLaST~\cite{chen2024llast}, our S2TT model consists of a speech encoder, an adaptor, and a large language model (LLM). For the TTS component, we adopt CosyVoice2~\cite{du2024cosyvoice}, a multilingual zero-shot TTS system designed for fine-grained prosody control, enabling precise modulation of emphasis in the output speech.

\begin{figure}[t!]
    \centering
        \includegraphics[width=0.49\textwidth]{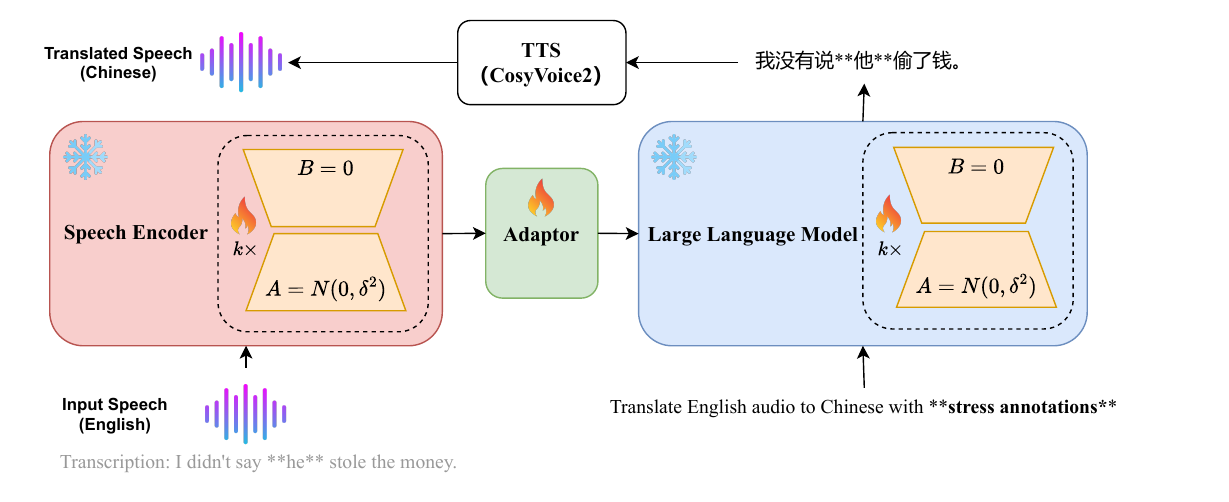} 
    \caption{Framework of the StressTransfer model.} 
    \label{fig:framework} 
\end{figure}

The workflow is structured as follows: Given source speech $\mathbf{S}$ in the source language, the speech encoder extracts acoustic and linguistic features, which are transformed by the adaptor to align with the LLM’s input space. The LLM, guided by a translation prompt, generates target-language text $Y_{tgt}$ interleaved with emphasis tags (e.g., \texttt{**word**}) that reflect stressed elements in the source speech. These tagged texts are then passed to the TTS module, which interprets the emphasis tags to modulate prosodic features such as pitch, energy, and duration, producing target speech that preserves the source’s emphatic intent. The whole framework is illustrated in Figure.~\ref{fig:framework}.

\subsection{Emphasis-Preserving S2TT Model}

Our S2TT model is designed to directly translate source speech into target-language text with embedded emphasis markers, capturing prosodic nuances end-to-end. The model comprises three core components, with a detailed focus on their architecture and interactions.

\noindent\textbf{Speech Encoder}: We employ Whisper-large-v3~\cite{radford2023robust} as the speech encoder $\mathcal{F}_{se}$, a transformer-based model pre-trained on large-scale multilingual speech data. The encoder takes raw audio $\mathbf{S}$ and extracts acoustic features $\mathbf{X}_s = \mathcal{F}_a(S)$ using a convolutional feature extraction layer that computes 128-channel Mel-spectrograms. These features are processed through transformer layers to produce linguistic representations $\mathbf{Z}_s = \mathcal{F}_{se}(\mathbf{X}_s)$. The output $\mathbf{Z}_s$ is a sequence of vectors capturing both phonetic and prosodic information, with a temporal resolution aligned to the input audio’s frame rate.

\noindent\textbf{Adaptor}: To bridge the modality gap between the speech encoder’s output and the LLM’s input space, we design a 3-layer multilayer perceptron (MLP) $\mathcal{F}_{ada}$ as the adaptor. In addition, we also performed downsampling to reduce the mismatch between the length of speech and text. The adaptor takes $\mathbf{Z}_s$ and projects it to an embedding space compatible with the LLM: $\mathbf{H}_s = \mathcal{F}_{ada}(\mathbf{Z}_s)$.

\noindent\textbf{Large Language Model (LLM)}: We use a decoder-only LLM, specifically a fine-tuned Qwen-2.5-3B~\cite{team2024qwen2}. The LLM takes concatenated inputs $[\mathbf{H}_s, \mathbf{H}_q]$, where $\mathbf{H}_q$ is the embedding of a text prompt $\textbf{X}_q$ (e.g., ``Translate to English, preserving emphasis with ** tags''). The prompt is tokenized using the LLM’s native tokenizer and embedded into the same dimensional space as $\mathbf{H}_s$. The LLM autoregressively generates the target text $\hat{Y}_{tgt} = \mathcal{F}_{llm}(\mathbf{H}_s, \mathbf{H}_q])$, interleaving emphasis tags where appropriate.

This architecture enables end-to-end learning of emphasis transfer, mapping source speech prosody directly to textual emphasis markers without requiring intermediate annotations during inference.

\subsection{Controllable TTS Module}

For speech synthesis, we adopt CosyVoice2~\cite{du2024cosyvoice}, an effective and advanced zero-shot TTS system designed for multilingual synthesis and prosody control. CosyVoice2 consists of a text encoder, a speech tokenizer for extracting supervised semantic tokens, an LLM for text-to-token generation, a conditional flow matching model for token-to-Mel-spectrogram conversion, and a HiFiGAN vocoder for waveform generation. The system supports fine-grained prosody control through text-based prompts, making it ideal for our emphasis-preserving pipeline. To encode emphasis, we post-process the S2TT output $\hat{Y}_{\text{tgt}}$ by converting Markdown \texttt{**...**} markers into CosyVoice2-compatible emphasis tags before using it as the custom prompt (e.g., \texttt{I didn't say <strong>he</strong> stole the money.}).

\subsection{Training and Inference}

\textbf{Training.} We fine-tune the S2TT component of StressTransfer on a combined dataset comprising standard S2TT data and our proposed EmphST-Instruct corpus. We employ the LoRA approach~\cite{chen2024llast}, applying it to both the speech encoder and the LLM, while the adaptor is fully fine-tuned. The training objective minimizes cross-entropy loss over emphasis-tagged target text. The TTS module remains unmodified, used in its original off-the-shelf configuration without fine-tuning. The details of the training data will be made public.

\textbf{Inference.} During inference, source speech $S$ is processed by the S2TT model to produce $\hat{Y}_{tgt}$ with emphasis tags. We use beam search (beam size=5) in the LLM to ensure high-quality text generation. The tagged text is then fed to CosyVoice2, which synthesizes the final speech with emphasis-modulated prosody. 

%% file: sec/sec5_experienments.tex
\section{Experienments}
In this section, we begin by describing the benchmark for evaluating the stress-aware speech-to-speech translation (S2ST) system, followed by an introduction to the experimental setup.
\vspace{-0.1in}
\subsection{EmphST-Bench}\label{sec:benchmark}
To guide algorithm exploration and evaluate the performance of our model, we design an evaluation pipeline for the emphasis-preserving speech-to-speech translation system. Given the lack of ready-to-use benchmarks for this important task, we leverage LLMs to translate the test set from the StressTest~\cite{yosha2025stresstest} corpus into the target language and then filter the results via human experts. This process creates a high-quality benchmark dataset, \textbf{EmphST-Bench}, with manually verified emphasis alignments between source and target utterances, ensuring reliable assessment of cross-lingual emphasis preservation. The human filtering step focuses on correcting any discrepancies in semantic equivalence, contrastive focus, and emotional intensity, resulting in a robust evaluation set that closely mirrors real-world linguistic nuances. EmphST-Bench consists of carefully selected parallel samples from English (source) to Chinese (target), providing a standardized resource for evaluating stress-aware S2ST systems. We report the statistics of EmphST-Bench in Table.~\ref{tab:emphst-bench-stats}.

\begin{table}[h]
\centering
\caption{Statistics of the EmphST-Bench dataset.}
\label{tab:emphst-bench-stats}
\resizebox{0.35\textwidth}{!}{%
\begin{tabular}{l|c}
\toprule
\textbf{Statistic} & \textbf{Value} \\
\midrule
Number of Samples & 218 \\
Avg. Audio Length (s) & 4.19 \\
Source Language & English \\
Target Language & Chinese \\
Avg. Source Text Length (words) & 6.95 \\
Avg. Target Text Length (character) & 10.87 \\
\bottomrule
\end{tabular}%
}

\end{table}
\vspace{-0.2in}
\subsection{Evaluation Methods}\label{sec:exp_config}
Our evaluation comprises two main parts: text-based and speech-based assessment.

\textbf{Text-based Evaluation.} We follow the Sentence Stress Reasoning Accuracy (SSR) ~\cite{yosha2025stresstest} metric to evaluate emphasis prediction in the translated text. SSR measures alignment between model-predicted emphasis and ground-truth annotations. For scalable evaluation, we treat emphasis recognition as a classification task and adopt an LLM-as-a-Judge framework, following common NLP practices for generative output evaluation\footnote{Due to space constraints, the detailed prompt will be provided in the full version.}. We also verify consistency between LLM and human judgments (see ablation study). Additionally, we report BLEU and COMET scores to assess semantic quality.
\textbf{Speech-based Evaluation.} We perform subjective human evaluations on synthesized speech to assess emphasis preservation. Native speakers rate the samples, providing validation for cross-lingual emphasis transfer. We also use ASR-BLEU and UTMOS~\cite{saeki2022utmos} to measure overall speech translation and audio quality.

\begin{table}[t!]
\small
\centering
\caption{Sentence Stress Reasoning Accuracy comparison of different methods on EmphST-Bench.}
\label{tab:stress_reasoning}
\resizebox{0.37\textwidth}{!}{
\begin{tabular}{clc}
\toprule \multirow{1}{*}{ID} & \multirow{1}{*}{Methods} & \multicolumn{1}{c}{SSR ( $\uparrow$ )} \\ 
\midrule \multicolumn{3}{c}{\textbf{SRC Text with Stress Prediction + LLM}} \\
\midrule
A1 & WhiStress + Qwen2.5-3B-Instruct &  72.9  \\

\midrule
\multicolumn{3}{c}{\textbf{Speech to Text Translation with Stress Prediction}} \\
\midrule A2 & GPT-4o & 47.4  \\
A3 & Gemini-2.5-Pro & 68.4   \\
A4 & StressTransfer &  \textbf{78.0} \\
\bottomrule
\end{tabular}
}

\end{table}

\subsection{Model Settings \& Baselines}
\textbf{Model Settings.} Our proposed model, StressTransfer, is a speech-to-speech translation system designed to preserve emphasis across languages. It is built upon a large language model architecture integrated with speech processing components. We use Whisper-Large-v3 as the speech encoder and Qwen2.5-3B as the LLM. The model is trained using LoRA techniques, and we employ the off-the-shelf CosyVoice2 for the TTS module.

 The model is trained on the constructed emphasis-aligned S2TT dataset \textbf{EmphST-Instruct} $\mathcal{D}_{\text{stress}}$ derived from the Stress17k and TinyStress corpus, with English-to-Chinese translation as the primary direction. We also used common public datasets such as CoVoST2~\cite{wang2021covost}, LibriSpeech~\cite{panayotov2015librispeech}, etc. We use the AdamW optimizer with a learning rate of 4e-5, a warm-up ratio of 0.03. Training is conducted 1 epoch or until convergence, with a batch size of 16 per GPU on 8 NVIDIA A100 GPUs. 

\textbf{Baseline Methods.} To assess the effectiveness of our text generation in the emphasis-preserving pipeline, we compare StressTransfer against two categories of baseline approaches: (a) end-to-end speech-to-text translation systems, such as GPT-4o or Gemini-2.5-Pro; and (b) cascaded systems that integrate ASR augmented with stress detection, exemplified by WhiStress~\cite{yosha2025whistress}, followed by machine translation using LLMs to incorporate the detected emphasis into the target text. For the speech-based evaluation, we compare our method not only with the aforementioned speech-to-text models combined with TTS synthesis but also with direct end-to-end speech-to-speech translation models(e.g. SeamlessExpressive~\cite{gong2024seamlessexpressivelm}).

%% file: sec/sec6_evaluation.tex
\section{Evaluation and Analysis}

\begin{table}[t!]
\small
\centering
\caption{Translation semantic quality evaluation results, including BLEU and COMET on CoVoST2 en2zh testset.}
\label{tab:semantic_quality}
\resizebox{0.38\textwidth}{!}{
\begin{tabular}{l|c|c}
\toprule  {Model} &{BLEU $\uparrow$ } & {COMET $\uparrow$}\\
\midrule SeamlessM4T-Large-V2~\cite{barrault2023seamless} & 35.9 & 0.8341 \\
SALMONN~\cite{tang2023salmonn} & 33.1 & - \\
Qwen2-Audio-7B~\cite{chu2024qwen2} & 45.2 & - \\

StressTransfer & \textbf{46.9} & \textbf{0.8347} \\
\bottomrule
\end{tabular}
}
\end{table}


\subsection{Evaluation on Generated Text}\label{sec:text_evaluation}

We first evaluate the systems' ability to accurately predict sentence stress in the generated text using the Sentence Stress Reasoning Accuracy (SSR)~\cite{yosha2025stresstest} metric on EmphST-Bench benchmark. Table~\ref{tab:stress_reasoning} presents the results, including SSR scores. Our proposed StressTransfer model demonstrates significant improvements in SSR compared to powerful proprietary LLMs, underscoring the effectiveness of our approach in preserving emphasis during cross-lingual translation. It is worth noting that the speech in EmphST-Bench is recording, while the audio source of $\mathcal{D}_{\text{stress}}$ is synthetic data, which does not overlap with the evaluation set.

In addition to stress preservation, we assess the semantic quality of the translations produced by different methods. Our method achieves competitive performance on the CoVoST-2 English-to-Chinese testset, as measured by BLEU and COMET\footnote{https://huggingface.co/Unbabel/wmt22-comet-da} scores, shown in Table~\ref{tab:semantic_quality}. The results show that the proposed method not only effectively preserves emphasis but also maintains highly competitive translation quality, with BLEU and COMET scores closely matching those of recent advanced speech translation models.

\subsection{Evaluation on Generated Speech}\label{sec:speech_evaluation}
For the synthesized speech outputs, we first conduct a subjective evaluation involving four models: the S2ST model focusing on prosody without LLM called SeamlessExpressive, the end-to-end GPT-4o-audio, which directly processes speech input to generate speech output; and two cascaded approaches—Gemini-2.5-Pro and our StressTransfer model—both integrated with an off-the-shelf TTS module. We enlisted four human experts to annotate the generated speech samples for emphasis preservation and compute the overall accuracy. The subjective results in Table~\ref{tab:asr_bleu} reveal that the cascaded system incorporating our proposed StressTransfer generates more effective emphasis-preserving translated speech, outperforming even top-tier proprietary models with only a small size of the parameters.

\begin{table}[htp!]
\small
\centering
\caption{ASR-BLEU Score, UTMOS and Subjective Sentence Stress Reasoning evaluation on generated speech.\label{tab:asr_bleu}}
\resizebox{0.45\textwidth}{!}{
\begin{tabular}{clccc}
\toprule \multirow{1}{*}{ID} & \multirow{1}{*}{\textbf{Methods}} & \multirow{1}{*}{\textbf{ASR-BLEU}($\uparrow$)} & \multirow{1}{*}{\textbf{UTMOS}($\uparrow$)}  & \multirow{1}{*}{\textbf{SSR}($\uparrow$)} \\
\midrule
C3 & GPT-4o-audio &  {49.38} & 3.46 & 41.3\\
C4 & Gemini-2.5-Pro + TTS & {48.47} & \textbf{3.91} & 45.0\\
\midrule C1 & SeamlessM4T-Large-V2 & {41.39} & 3.52 & -\\
C2 & SeamlessExpressive & {35.06} & 2.63 & 11.0\\
\midrule
C5 & StressTransfer + TTS &  \textbf{50.26} & 3.86  & \textbf{46.0}\\
\bottomrule
\end{tabular}
}
\end{table}

\vspace{-0.05in}

Additionally, we objectively evaluate translation quality using the ASR-BLEU metric on the StressTest corpus, which involves transcribing output speech and computing ASR-BLEU scores against reference texts. Table~\ref{tab:asr_bleu} presents the results, where our model achieves the competitive ASR-BLEU score, even compared with the proprietary models, GPT-4o-Audio and the Gemini-2.5-Pro + TTS pipeline.
SeamlessM4T-Large-V2 and SeamlessExpressive demonstrate a noticeable performance gap in ASR-BLEU scores compared to the other three systems. This discrepancy likely stems from inherent limitations in Chinese language modeling within these frameworks. SeamlessM4T-Large-V2, while massively multilingual, exhibits reduced optimization for tonal languages like Chinese where phonological distinctions critically impact semantic meaning. The challenge is amplified in SeamlessExpressive, evidenced by occasional intelligibility issues in generated Chinese utterances. 

Furthermore, we employ the UTMOS metric to evaluate the audio quality of each system's output. As shown in Table~\ref{tab:asr_bleu}, both cascade-based models achieve the highest UTMOS scores, demonstrating the superiority of our system. Notably, these two systems utilize identical TTS components, resulting in marginally higher scores compared to other approaches. Conversely, SeamlessExpressive registers the lowest score, aligning with our previous observation regarding its suboptimal Chinese speech generation quality. Audio samples are available at \footnote{https://chenx17.github.io/demos/stresstrans/demo.html}.


\subsection{Ablation Study}\label{sec:ablation}
We further conduct ablation study to investigate the influence of the training dataset and seasonality LLM-as-Judge, also we conduct the detailed error mode analysis of the StressTransfer.

\textbf{Influence of the Data Recipe.}
We conduct an ablation study by training the model with and without the EmphST-Instruct dataset. As shown in Table~\ref{tab:ablation_data}, although models trained without EmphST-Instruct achieve slightly higher BLEU and COMET scores on the CoVoST2 English-to-Chinese test set, the difference is marginal. More importantly, incorporating EmphST-Instruct enables stress-aware speech translation, validating its crucial role in achieving emphasis-preserving translation.

\textbf{Reliability of LLM-as-Judge.}\label{sec:llm_as_judge}
To quantify the alignment between the \textbf{LLM-as-Judge} (GPT-4.1) and human raters, we carried out a consistency analysis on predictions produced by two underlying systems—\textbf{Gemini-2.5 Pro} and \textbf{StressTransfer}. Seven native human testers independently annotated the emphasized spans in each prediction. Their labels were merged by majority vote to create a single reference per instance, against which the GPT-4.1 verdicts were compared. The resulting agreement scores are reported in Table~\ref{tab:consistency}. As the table shows, GPT-4.1 achieves a stable and satisfactory level of agreement with the expert panel across both models( 0.9271 on F1 score), confirming its reliability on this emphasis-reasoning task.

\begin{table}[t!]
\small
\centering
\caption{Ablation study on training data for StressTransfer}
\label{tab:ablation_data}
\resizebox{0.38\textwidth}{!}{
\begin{tabular}{l|c|c}
\toprule  {\textbf{Data}} &{\textbf{BLEU} ($\uparrow$) } & {\textbf{COMET} ($\uparrow$)}\\
\midrule
Full Data & 46.9 & 0.8347 \\
w/o EmphST-Instruct & 47.4 & 0.8362 \\
\bottomrule
\end{tabular}
}
\end{table}


\begin{table}[t]
  \centering
\caption{Overall consistency metrics for GPT-4.1 as judge compared with the expert consensus.}
\label{tab:consistency}
\resizebox{0.38\textwidth}{!}{
  \begin{tabular}{lcccc}
    \toprule
    \textbf{Metric} & \textbf{Accuracy} & \textbf{Precision} & \textbf{Recall} & \textbf{F1} \\
    \midrule
    \textbf{Score} & 0.8704 & 0.9219 & 0.9323 & 0.9271 \\
    \bottomrule
  \end{tabular}}
\end{table}
